\documentclass[letterpaper, 10 pt, conference]{ieeeconf}  % Comment this line out if you need a4paper

\IEEEoverridecommandlockouts                              % This command is only needed if 
\title{\LARGE \bf
PhaForce: Phase-Scheduled Visual–Force Policy Learning with Slow Planning and Fast Correction for Contact-Rich Manipulation
}

\author{
Mingxin Wang, Zhirun Yue, Renhao Lu, Yizhe Li, Zihan Wang,\\
Guoping Pan, Kangkang Dong, Jun Cheng, Yi Cheng, Houde Liu$^{*}$
}

\usepackage{cite}
\usepackage{graphicx}
\usepackage{amsmath}
\usepackage{amssymb}
\usepackage{amsfonts}
\usepackage{booktabs}
\usepackage{tabularx}
\usepackage{pifont}
\usepackage{hyperref}

\newcommand{\cmark}{\ding{51}}

\begin{document}

\maketitle

\thispagestyle{empty}
\pagestyle{empty}

%%%%%%%%%%%%%%%%%%%%%%%%%%%%%%%%%%%%%%%%%%%%%%%%%%%%%%%%%%%%%%%%%%%%%%%%%%%%%%%%
\begin{abstract}
Contact-rich manipulation requires not only vision-dominant task semantics but also closed-loop reactions to force/torque (F/T) transients.
Yet, generative visuomotor policies are typically constrained to low-frequency updates due to inference latency and action chunking, underutilizing F/T for control-rate feedback.
Furthermore, existing force-aware methods often inject force continuously and indiscriminately, lacking an explicit mechanism to schedule \emph{when / how much / where} to apply force across different task phases.
We propose \textbf{PhaForce}, a phase-scheduled visual--force policy that coordinates low-rate chunk-level planning and high-rate residual correction via a unified \emph{contact/phase} schedule.
PhaForce comprises (i) a contact-aware phase predictor (CAP) that estimates contact probability and phase belief, (ii) a Slow diffusion planner that performs dual-gated visual--force fusion with \emph{orthogonal residual injection} to preserve vision semantics while conditioning on force, and (iii) a Fast corrector that applies \emph{control-rate} phase-routed residuals in interpretable corrective subspaces for within-chunk micro-adjustments.
Across multiple real-robot contact-rich tasks, PhaForce achieves an average success rate of 86\% (+40 pp over baselines), while also substantially improving contact quality by regulating interaction forces and exhibiting robust adaptability to OOD geometric shifts.
\end{abstract}

%%%%%%%%%%%%%%%%%%%%%%%%%%%%%%%%%%%%%%%%%%%%%%%%%%%%%%%%%%%%%%%%%%%%%%%%%%%%%%%%
\section{INTRODUCTION}

% 第一段：视觉IL/VLA取得的进展和富接触任务的难点
Diffusion-based visuomotor policies~\cite{chi2025diffusion} and recent VLA models~\cite{kim2024openvla,black2024pi_0,intelligence2025pi_} have achieved strong performance on vision-dominant manipulation tasks such as pick-and-place, rearrangement, and folding~\cite{zhao2023learning,liu2024rdt}.
However, many real-world skills are inherently \emph{contact-rich}: success depends not only on geometric alignment but also on interaction dynamics such as friction, jamming, and transient impacts~\cite{sun2024force,stepputtis2022system,zhang2026dextac,wu2025tacdiffusion,tsuji2024adaptive,chen2025dexforce}.
In such scenarios, vision is often ambiguous or occluded, and critical signals emerge as short-horizon 6D force/torque (F/T) events.
For instance, in insertion, being fully seated versus being jammed on the rim can be visually indistinguishable at millimeter scale, while wrench transients reveal misalignment and recovery cues~\cite{zhang2025ta}.
Similarly, in wiping, visual observations rarely reveal whether the tool is slightly detached or over-pressed~\cite{tsuji2024adaptive}.

% 第三段：现在融入force的IL/VLA是怎么做的
This motivates incorporating F/T (wrench) sensing as physical feedback for contact-rich manipulation.
Most force-aware policies encode a short F/T history and fuse it with vision (e.g., concatenation or attention), then use the fused multimodal representation in a chunked generative policy~\cite{hou2025adaptive,li2025adaptive,yu2025forcevla,zhou2025admittance}.

%第四段：现在工作的缺陷
However, a key structural mismatch remains underexplored \textbf{(Gap-1: timescale mismatch of force feedback)}: F/T is a feedback signal whose value lies in rapid closed-loop correction, while generative policies are typically constrained to low-frequency updates by inference latency and action chunking.
When force is primarily consumed at the action-chunk update rate, short-horizon interaction transients (e.g., stick--slip, micro-impacts, early jamming) can be under-reacted.
This calls for an explicit closed-loop correction layer that reacts to force feedback within an action chunk.

\begin{figure}[t]
  \centering
  % 调整 trim={<left> <bottom> <right> <top>} 的数值来裁剪图片，单位是bp
  \includegraphics[width=1\columnwidth, trim={30 0 0 0}, clip]{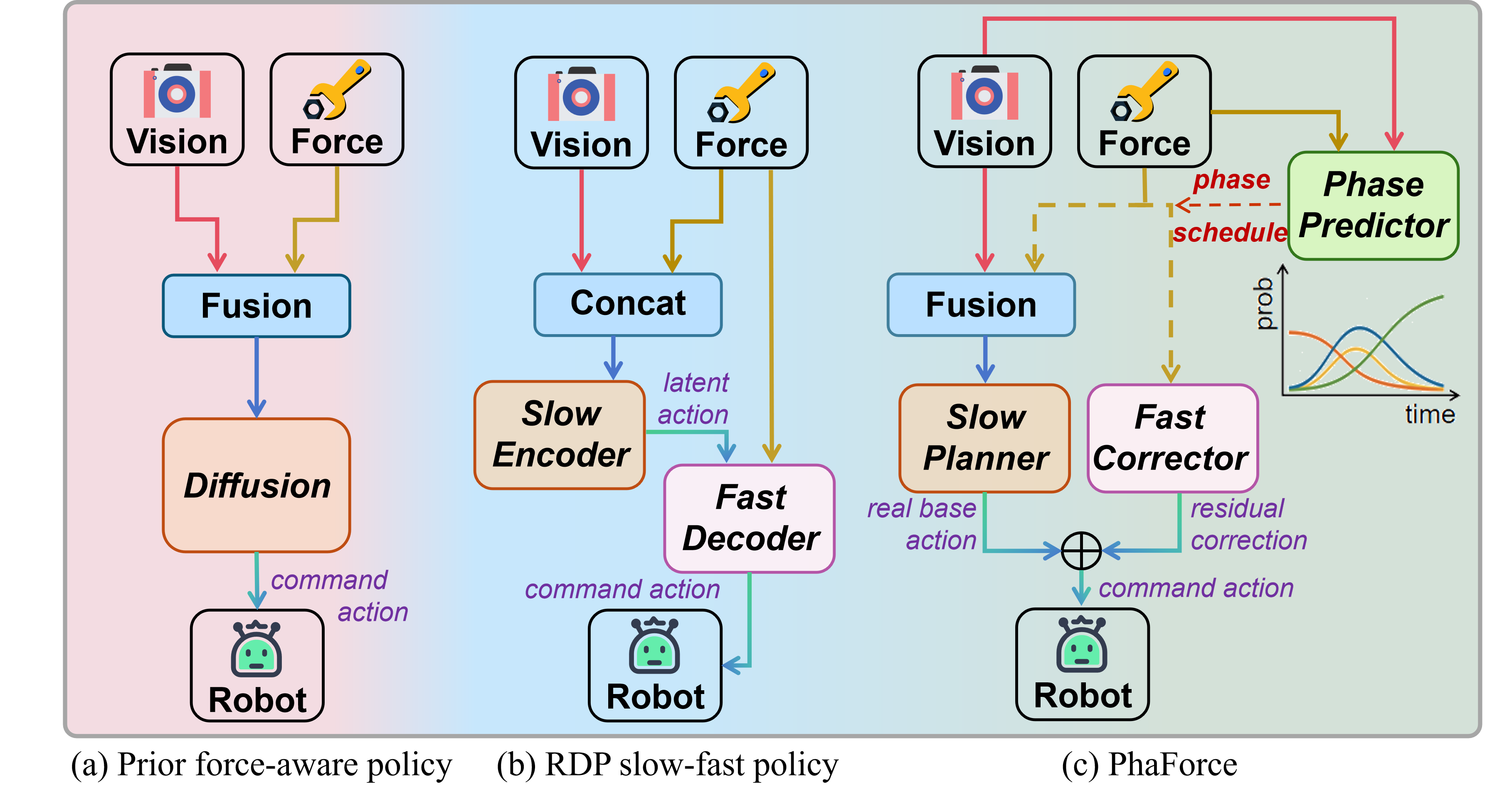}
    \caption{
    Comparison of three force-aware policy architectures.
    Prior works fuse vision and force into a single generative policy, while RDP adopts a slow--fast decomposition without explicit phase scheduling.
    \textbf{PhaForce} introduces an explicit contact/phase schedule to coordinate force usage for both chunk-level planning (Slow) and within-chunk correction (Fast).
    }
  \label{fig:intro}
\end{figure}

%第五段：引出RDP的重要性和其不足之处
Reactive Diffusion Policy~\cite{xue2025reactive} takes an important step towards slow--fast execution by coupling chunk-level generative planning with within-chunk reactivity.
However, such reactive designs remain largely \emph{phase-agnostic}, often applying high-frequency corrections without explicitly distinguishing which motion channels should be corrected at different stages.
Contact-rich manipulation is inherently multi-phase: different stages (e.g., planar search versus normal insertion) demand orthogonal or even mutually exclusive corrective subspaces.
Without an explicit phase schedule, high-rate reactivity can introduce spurious corrections in irrelevant subspaces, degrading alignment and potentially leading to jamming behaviors.
Prior work therefore lacks an explicit phase schedule to dynamically route force feedback---leaving a critical gap in deciding \emph{when} to trust force, \emph{how much} to use it, and critically, \emph{where} (in which corrective subspace) to apply it \textbf{(Gap-2: explicit phase scheduling)}~\cite{he2025foar,lee2025manipforce}.

%第六段：本篇工作的方法概述
In this work, we propose \textbf{PhaForce}, a \emph{phase-scheduled} slow--fast policy that uses an explicit contact probability and a task-defined phase-belief distribution to coordinate force usage for both chunk-level planning and within-chunk closed-loop correction.
Fig.~\ref{fig:intro} provides an intuitive comparison of three force-aware policy architectures.
PhaForce consists of three components:
(1) a \emph{contact-aware phase predictor} that outputs a continuous contact probability and a soft distribution over phases, providing an explicit semantic schedule signal;
(2) a \textbf{Slow} diffusion planner that performs \emph{dual-gated} visual--force fusion for long-horizon action-chunk generation, where contact gates the overall force injection and phase belief modulates the fused representation to maintain vision-dominant semantics via \emph{orthogonal residual injection};
and (3) a \textbf{Fast} residual corrector that applies control-rate corrections in \emph{phase-routed} corrective subspaces, trained with physically motivated supervision constructed from the \emph{virtual target pose}.
The final control command is obtained by composing the \textbf{Slow} real base action with the \textbf{Fast} residual correction.

%第七段：三个contribution
Our contributions are summarized as follows:
\begin{itemize}
  \item We propose \textbf{PhaForce}, a \emph{phase-scheduled} slow--fast policy that unifies force-aware chunk-level generative planning with control-rate residual correction.
  \item We introduce an explicit scheduling signal (contact probability + phase belief) that decides \emph{when / how much} to use force for planning and \emph{where} to correct during execution, realized by dual-gated fusion with orthogonal residual injection in \textbf{Slow} and phase-routed corrective subspaces in \textbf{Fast}.
  \item We validate PhaForce on multiple real-robot contact-rich manipulation tasks, showing consistent improvements over strong baselines in both ID and OOD settings.
\end{itemize}

\section{RELATED WORK}

\subsection{Force-Aware Visuomotor Policy Learning}

A prevalent paradigm for force-aware imitation learning encodes a short history window of 6D wrench measurements and fuses the resulting force representation with visual features—typically via feature concatenation~\cite{hou2025adaptive,zhou2025admittance,li2025flow,liu2025forcemimic} or attention-based cross-modal interaction~\cite{li2025adaptive,ge2025filic,choi2026wild,aburub2024learning,kamijo2024learning,kang2025robotic}, treating the multimodal feature as the observation for a visuomotor policy.
FoAR~\cite{he2025foar} modulates force usage with a predicted future contact probability to suppress noisy wrench signals in free-space and amplify force under contact.
ForceVLA~\cite{yu2025forcevla} employs a force-aware Mixture-of-Experts block, where expert routing varies with task progress and can implicitly specialize across interaction stages.

Beyond fusion, Stepputtis et al.~\cite{stepputtis2022system} introduce a continuous phase variable (from 0 to 1) to represent task progress and feed it to skill primitives for contact-rich manipulation.
TA-VLA~\cite{zhang2025ta} highlights torque transients as reliable event signals that reveal contact outcomes and naturally support intent switching (e.g., detect failure and retry). 

However, existing approaches still lack an explicit, task-defined \emph{probabilistic phase belief} mechanism to interpretably schedule \emph{when / how much} force should be fused with vision and to exploit force cues without degrading visual task semantics.

\subsection{Slow--Fast Policy Learning under Action Chunking}

Despite recent progress in force-aware visuomotor policy learning, many methods still rely on action-chunk generation, resulting in near open-loop execution within each chunk and delaying the use of wrench transients that often signal contact anomalies and intent switches.
ManipForce~\cite{lee2025manipforce} introduces frequency-aware multimodal representations, but still follows the chunked diffusion paradigm without dedicated within-chunk force-driven correction.

Reactive Diffusion Policy~\cite{xue2025reactive} adopts a slow--fast architecture, where a slow diffusion model predicts low-rate latent action chunks and a fast decoder leverages high-rate wrench feedback to autoregressively generate fine-grained control commands within each chunk.
Subsequent analyses~\cite{chen2025implicitrdp} suggest that latent chunk compression may reduce free-space motion precision and degrade millimeter-level approach/contact localization.
More broadly, wrench feedback in existing designs is still primarily used for within-chunk local refinement, leaving open a structured, task-semantic mechanism for long-horizon intent switching and phase-dependent correction.
ImplicitRDP~\cite{chen2025implicitrdp} further explores end-to-end structural slow--fast learning, yet explicit and interpretable phase-level scheduling remains underexplored.

Overall, it remains underexplored how to close the loop at high control rates while coordinating long-horizon planning and residual correction in an interpretable, phase-dependent manner.

\begin{figure*}[!t]
    \centering % 图片居中
    % 调整width=1\textwidth使图片铺满双栏宽度
    \includegraphics[width=1\textwidth]{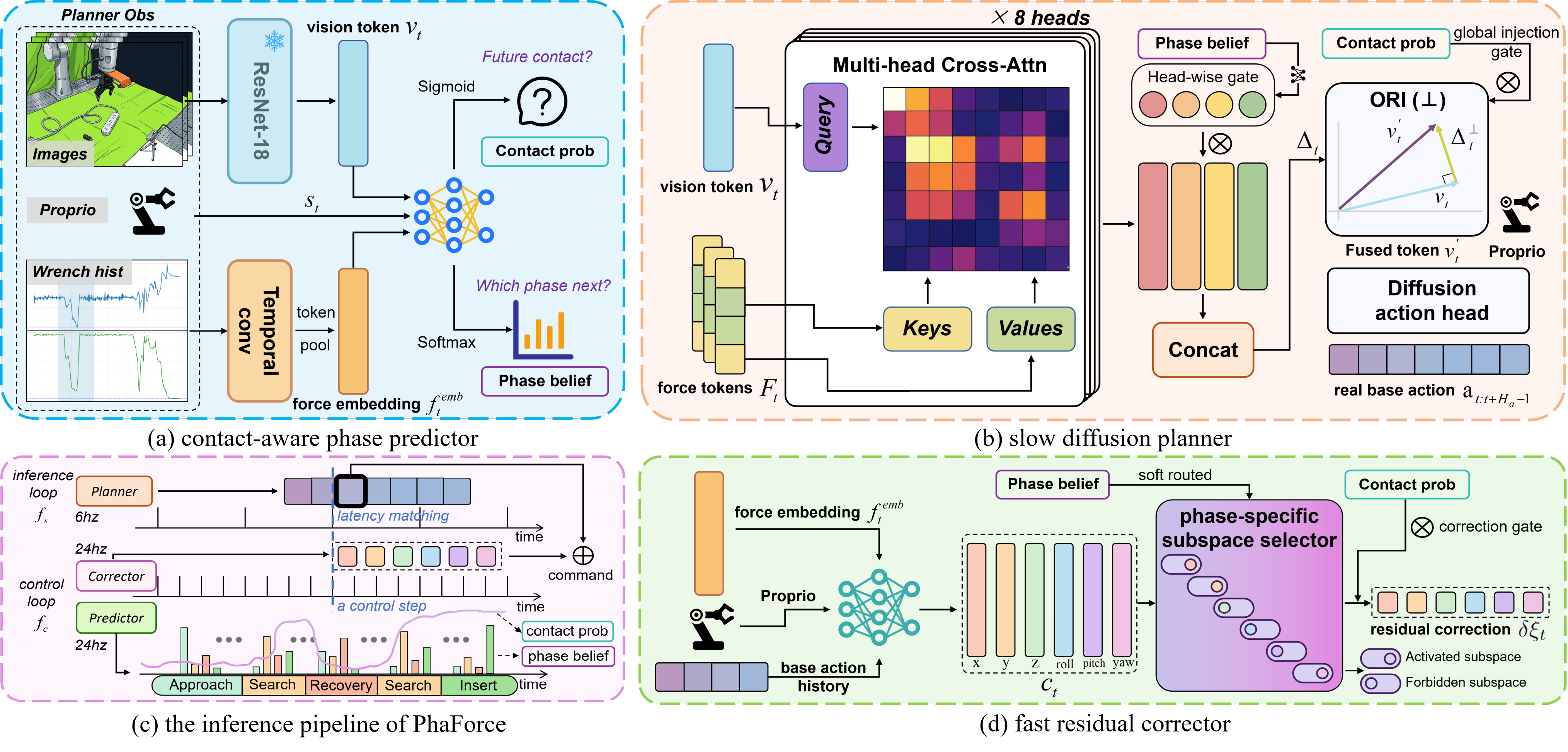}
    % 图片标题，label用于交叉引用
     \caption{\textbf{PhaForce Architecture.} 
    The \textbf{Slow} diffusion planner runs at $f_s{=}6$~Hz to generate action chunks, while \textbf{CAP} and the \textbf{Fast} corrector run at the control rate $f_c{=}24$~Hz for contact/phase prediction and within-chunk closed-loop correction.
    In \textbf{Slow}, dual-gated vision--force fusion with orthogonal residual injection preserves vision-dominant task semantics.
    In \textbf{Fast}, phase-belief soft routing activates corrective subspaces and outputs a residual correction that is composed with the Slow base action to obtain the executed command.}
     
    \label{fig:pipeline}
\end{figure*}

\section{METHOD}

In this section, we present \textbf{PhaForce}, a phase-scheduled visual--force policy for contact-rich manipulation (Fig.~\ref{fig:pipeline}).
We first formalize the problem and specify the slow--fast execution pipeline under action chunking (Sec.~\ref{sec:prelim}).
We then introduce a \emph{contact-aware phase predictor} that outputs a continuous contact probability and a phase-belief distribution (Sec.~\ref{sec:cap}).
Finally, we describe how this phase belief coordinates both low-frequency task-intent planning in the \textbf{Slow} planner (Sec.~\ref{sec:slow}) and high-frequency residual correction in the \textbf{Fast} corrector (Sec.~\ref{sec:fast}).

\subsection{Problem Formulation and Preliminaries}
\label{sec:prelim}

\textbf{Observation inputs.}
At each timestep $t$, we define two observation views tailored to slow--fast execution.
The \emph{planner observation} $o_t^{p}=(\mathcal{I}_t,\; w_t^{\mathrm{hist}},\; s_t)$ includes multi-view RGB images $\mathcal{I}_t$, wrench history $w_t^{\mathrm{hist}}$ expressed in the TCP frame, and proprioception $s_t$.
The \emph{corrector observation} $o_t^{c}=(w_t^{\mathrm{hist}},\; s_t)$ excludes images and uses only low-dimensional signals.
Here we define $H_w$ as the wrench-history window length, capturing short-term interaction dynamics.

\textbf{Policy outputs.}
We learn a slow--fast policy pair $(\pi_{\text{slow}},\pi_{\text{fast}})$ under action chunking with two update rates.
As shown in Fig.~\ref{fig:pipeline}(c), $f_c$ denotes the \emph{control frequency} at which the robot is commanded and the \textbf{Fast} corrector is evaluated,
and $f_s$ denotes the \emph{inference frequency} of the \textbf{Slow} planner, typically $f_s \ll f_c$ due to inference latency.
$\pi_{\text{slow}}$ predicts a nominal action chunk; we denote by
$T_t^{\text{slow}}\in SE(3)$ the corresponding nominal TCP pose at control step $t$.
In contrast, $\pi_{\text{fast}}$ consumes $o_t^{c}$ at $f_c$ and predicts a small delta pose $\Delta T_t^{\text{fast}}\in SE(3)$ in the TCP frame for high-rate residual correction.
The executed pose is composed on $SE(3)$ as
\begin{equation}
T_t = T_t^{\text{slow}} \circ \Delta T_t^{\text{fast}}.
\label{eq:se3_comp}
\end{equation}
In practice, we represent each pose $T_t\in SE(3)$ by position and a unit quaternion, and send the low-level command
$a_t\in\mathbb{R}^8$ with gripper width.

\subsection{Contact-Aware Phase Predictor}
\label{sec:cap}

Beyond contact state, contact-rich tasks are inherently multi-phase; different phases demand different force usage and corrective subspaces (e.g., planar search vs.\ normal compliance).
To explicitly represent such task progress, for each task we define $K$ task-specific phases (e.g., \emph{approach/search/recovery/insert/done} in plug-in tasks) and predict a continuous contact probability $p_t^{c}\in[0,1]$ and a phase belief $\mathbf{p}_t\in\Delta^{K-1}$ (with $\sum_{k=1}^{K}\mathbf{p}_t^{(k)}=1$), which are evaluated at $f_c$ and used to schedule force usage in \textbf{Slow} and \textbf{Fast} (Secs.~\ref{sec:slow}--\ref{sec:fast}).

\textbf{Inputs and outputs.}
As shown in Fig.~\ref{fig:pipeline}(a), we introduce a lightweight predictor \textbf{CAP}, denoted as $\pi_{\mathrm{CAP}}$, which takes the planner observation $o_t^{p}=(\mathcal{I}_t, w_t^{\mathrm{hist}}, s_t)$ as input and outputs $(p_t^{c}, \mathbf{p}_t)$.
We use a ResNet-18 for each RGB view without weight sharing to extract visual features, which are fused with force/proprio features by a small MLP, followed by a binary contact head and a categorical phase head.

\textbf{Force encoder.}
To encode the wrench history $w_t^{\mathrm{hist}}$ while capturing both abrupt interaction transients and short-term temporal dependencies, we use a lightweight TCN-style temporal encoder shared across \textbf{CAP}, \textbf{Slow}, and \textbf{Fast}.
It is implemented as a stack of dilated 1D convolutions with residual connections, providing multi-scale temporal receptive fields that are sensitive to short-lived F/T changes.
For \textbf{Slow}, the encoder outputs a sequence of force tokens $\{f_{t-i}\}_{i=0}^{H_w-1}$ with $f_{t-i}\in\mathbb{R}^{d}$; for modules that require a single vector (\textbf{CAP}/\textbf{Fast}), we additionally apply temporal pooling to obtain a compact force embedding $f_t^{\mathrm{emb}}\in\mathbb{R}^{d_f}$.

\textbf{Targets and loss.}
Importantly, $\pi_{\mathrm{CAP}}$ is trained for \emph{anticipation} rather than instantaneous judgment.
We supervise the contact head using a future-window label indicating whether contact will occur within the next $K_f$ control steps:
$y_t^{c}=\bigvee_{i=1}^{K_f}\mathrm{contact}_{t+i}$
and supervise the phase head using a future offset label $y_t^{\text{phase}}=\mathrm{phase}_{t+\delta}$.
Let $\ell_t^{c}\in\mathbb{R}$ and $\ell_t^{\phi}\in\mathbb{R}^{K}$ denote the contact/phase logits, with $p_t^{c}=\sigma(\ell_t^{c})$ and $\mathbf{p}_t=\mathrm{softmax}(\ell_t^{\phi})$.
We optimize a multi-task objective:
\begin{equation}
\mathcal{L}_{\mathrm{CAP}}
=
\mathcal{L}_{\mathrm{BCE}}(y_t^{c}, \ell_t^{c})
+
\lambda_{\phi}\,\mathcal{L}_{\mathrm{CE}}(y_t^{\text{phase}}, \ell_t^{\phi}),
\label{eq:cap_loss}
\end{equation}
where $\mathcal{L}_{\mathrm{BCE}}$ is binary cross-entropy (with logits) for future contact prediction and $\mathcal{L}_{\mathrm{CE}}$ is cross-entropy for phase prediction.
All labels are automatically generated via scripts using wrench signals and TCP pose, avoiding manual annotation.

\subsection{Slow Diffusion Planner}
\label{sec:slow}

As shown in Fig.~\ref{fig:pipeline}(b), the \textbf{Slow} diffusion planner $\pi_{\mathrm{slow}}$ runs at rate $f_s$ with an augmented planner input
$\tilde{o}_t^{p}=(o_t^{p},\,p_t^{c},\,\mathbf{p}_t)$ and outputs an executable action chunk of horizon $H_a$ in control steps:
\begin{equation}
\mathbf{a}_{t:t+H_a-1}\sim \pi_{\mathrm{slow}}(\,\cdot \mid \tilde{o}_t^{p}\,).
\end{equation}

\textbf{Encoders.}
We encode the multi-view RGB observation into a single visual token $v_t\in\mathbb{R}^{d}$ by concatenating per-view global embeddings extracted by ResNet-18 encoders,
and encode the wrench history into force tokens $F_t\in\mathbb{R}^{H_w\times d}$ using the force encoder described in Sec.~\ref{sec:cap}.

\textbf{Dual-gated fusion.}
We fuse vision and force via a multi-head cross-attention block, which uses the visual token as a query to attentively aggregate the force tokens.

Let $Q=v_t W_Q\in\mathbb{R}^{1\times d_k}$ and $K=F_t W_K,\ V=F_t W_V\in\mathbb{R}^{H_w\times d_k}$ where $W_Q,W_K,W_V\in\mathbb{R}^{d\times d_k}$.
For a single head,
\begin{equation}
\mathrm{Attn}(Q,K,V)=\mathrm{softmax}\!\left(\frac{QK^\top}{\sqrt{d_k}}\right)V.
\label{eq:slow_attn}
\end{equation}

To make force usage phase-dependent and interpretable, we introduce a phase-dependent \emph{head-wise gate}
\begin{equation}
g_t^{\mathrm{head}}=\sigma\!\left(\mathrm{MLP}(\mathbf{p}_t)\right)\in[0,1]^H,
\label{eq:slow_head_gate}
\end{equation}
where $g_t^{\mathrm{head}}(h)$ denotes its $h$-th element.
We reweight per-head outputs by $g_t^{\mathrm{head}}$ and obtain the cross-attention output
\begin{equation}
\Delta_t
=
W_O\big[\,g_t^{\mathrm{head}}(1)\,\mathrm{Attn}_1;\ \ldots;\ g_t^{\mathrm{head}}(H)\,\mathrm{Attn}_H\,\big],
\label{eq:slow_mh_attn}
\end{equation}
where $\mathrm{Attn}_h$ denotes the output of head $h$ computed by Eq.~\eqref{eq:slow_attn} with head-specific projections, and $W_O$ is the output projection.
In addition, a \emph{global injection gate} $g_t^{c}=p_t^{c}$ controls the injection strength in Eq.~\eqref{eq:slow_fused}, suppressing the influence of noisy wrench signals in free-space.

\textbf{Orthogonal residual injection (\emph{ORI}).}
Rather than overwriting the visual feature with $\Delta_t$, we inject it as a \emph{residual} and retain only its component \emph{orthogonal} to the visual token, preserving vision-dominant semantics and mitigating semantic drift.

\begin{equation}
\Delta_t^{\perp}
=
\Delta_t-\mathrm{Proj}_{v_t}(\Delta_t)
=
\Delta_t-\frac{\langle \Delta_t,v_t\rangle}{\langle v_t,v_t\rangle+\epsilon}\,v_t,
\label{eq:slow_orth}
\end{equation}

where $\epsilon=10^{-6}$ is a numerical stabilizer.
The fused token is then
\begin{equation}
v_t' = v_t + \alpha\cdot g_t^{c}\cdot \Delta_t^{\perp},
\label{eq:slow_fused}
\end{equation}
where $\alpha$ is a learnable scalar gain that is clipped to a bounded range for stability. Intuitively, $p_t^{c}$ controls \emph{when / how much} force should influence planning, while $\mathbf{p}_t$ controls \emph{which heads} are emphasized under the current phase.

\textbf{Diffusion-based chunk planning.}
Given the conditioning $z_t=\mathrm{cond}(v_t',\,s_t)$, we use a diffusion action head to generate an action chunk by progressively denoising a noisy action trajectory into executable commands~\cite{chi2025diffusion,ho2020denoising,song2020denoising}. In training, the rotational component is represented by 6DRot.

% =======================
% METHOD Sec.4: Fast (compact, ~half page)
% =======================
\subsection{Fast Residual Corrector}
\label{sec:fast}

As shown in Fig.~\ref{fig:pipeline}(d), the \textbf{Fast} corrector $\pi_{\mathrm{fast}}$ runs at rate $f_c$ with an augmented corrector input
$\tilde{o}_t^{c}=(\,o_t^{c},\ \mathbf{h}_t^{\mathrm{slow}},\ p_t^{c},\ \mathbf{p}_t\,)$,
where $\mathbf{h}_t^{\mathrm{slow}}$ denotes a short history of base actions produced by \textbf{Slow}.
\textbf{Fast} predicts an intermediate within-chunk \emph{channel-wise} residual $c_t$:
\begin{equation}
c_t = \pi_{\mathrm{fast}}(\tilde{o}_t^{c})\in\mathbb{R}^{6},
\label{eq:fast_c}
\end{equation}
where $c_t=\left[c_x,\ c_y,\ c_z,\ c_{\mathrm{roll}},\ c_{\mathrm{pitch}},\ c_{\mathrm{yaw}}\right]^{\top}$.

\textbf{Phase-routed corrective subspaces.}
The channel residual $c_t$ specifies residual increments along interpretable correction channels.
For each phase $k\in\{1,\ldots,K\}$, we predefine a \emph{phase-specific subspace selector}
$B_k\in\mathbb{R}^{6\times 6}$ as a diagonal binary channel mask, i.e., $B_k=\mathrm{diag}(m_k)$ with $m_k\in\{0,1\}^{6}$, which disables \emph{forbidden} dimensions and keeps only the \emph{activated} channels in that phase.
For example, in plug-in tasks, the \emph{search} phase activates $(x, y, \text{yaw})$
(e.g., $m_{\mathrm{search}}=[1,1,0,0,0,1]$),
whereas the \emph{insert} phase activates normal compliance along $(z, \text{yaw})$.
We then obtain the residual twist by \emph{softly routing} the raw channel-wise residual $c_t$ using the phase belief, with contact gating by $p_t^{c}$:
\begin{equation}
\delta\xi_t
=
p_t^{c}\left(\sum_{k=1}^{K}\mathbf{p}_t^{(k)}\,B_k\right)c_t
\in\mathbb{R}^{6},
\label{eq:fast_delta_xi}
\end{equation}
For execution, $\delta \xi_t$ is converted to the pose increment $\Delta T_t^{\mathrm{fast}}$. Unlike admittance controllers that typically rely on fixed gains and hand-designed DOF switching, phase-belief soft routing smoothly interpolates among phase-specific corrective subspaces.
Moreover, contact gating suppresses spurious corrections induced by free-space wrench noise without additional heuristic thresholds or filtering.

\textbf{Physical-prior supervision.}
In contact-rich tasks, beyond the nominal TCP pose, we consider a \emph{virtual target pose} that the robot would track under compliant interaction, as implied by force feedback~\cite{hou2025adaptive}.
Rather than explicitly estimating this target pose, we treat the desired \emph{pose offset} as a residual twist and specify it via a phase-dependent physical prior, yielding automatic supervision for \textbf{Fast}.
Concretely, we construct phase-wise physically motivated residual targets $\delta\xi^{\ast}_{t,k}\in\mathbb{R}^{6}$ from wrench signals to capture desired corrective trends in each phase.
For example, during planar \emph{search}, the target drives the robot to relieve tangential friction and mitigate jamming torques:
\begin{equation}
\delta\xi^{\ast}_{t,\mathrm{search}}
=
\left[-\alpha_{x}F_{x,t},\ -\alpha_{y}F_{y,t},\ 0,\ 0,\ 0,\ -\alpha_{\mathrm{yaw}}\tau_{z,t}\right]^{\top},
\label{eq:fast_teacher_search}
\end{equation}
where we treat roll/pitch as a \emph{forbidden subspace} to promote stable execution. During \emph{wiping}, the target enforces normal compliance by tracking a desired normal force $F_z^{\ast}$:
\begin{equation}
\delta\xi^{\ast}_{t,\mathrm{wiping}}
=
\left[0,\ 0,\ \alpha_z\!\left(F_z^{\ast}-F_{z,t}\right),\ 0,\ 0,\ 0\right]^{\top}.
\label{eq:fast_teacher_wiping}
\end{equation}
Similar targets can be defined for rotation channels using measured torques.
To keep supervision consistent with the soft-routed correction in Eq.~\eqref{eq:fast_delta_xi}, we compute a single target by phase-belief--weighted averaging of phase-wise residual priors, with correction gating by $p_t^{c}$:
\begin{equation}
\delta\xi^{\ast}_t
=
p_t^{c}\sum_{k=1}^{K}\mathbf{p}_t^{(k)}\,\delta\xi^{\ast}_{t,k}.
\label{eq:fast_teacher_mix}
\end{equation}

\textbf{Training loss.}
We regress $\delta\xi_t$ to $\delta\xi^{\ast}_t$ with an $\ell_1$ loss:
\begin{equation}
\mathcal{L}_{\mathrm{fast}}
=
\mathbb{E}\left[\left\|\delta\xi_t-\delta\xi^{\ast}_t\right\|_{1}\right].
\label{eq:fast_loss}
\end{equation}

\section{EXPERIMENTS}

\subsection{Experimental Setup}
Our experiments are conducted on a Flexiv Rizon~4s robotic arm equipped with a 6-axis force/torque sensor at the end effector.
We use one wrist-mounted and two external Intel RealSense D435 cameras to provide multi-view RGB observations.
We collect 80 expert teleoperated demonstrations per task using TactAR~\cite{xue2025reactive}, which provides real-time \emph{wrench visualization} to refine contact behaviors.
All devices are connected to a workstation with an Intel Core i7-14700F CPU and an NVIDIA RTX 4060~Ti GPU for data collection and policy evaluation.

\subsection{Tasks and Metrics}

\textbf{Tasks.}
We evaluate PhaForce on five real-robot contact-rich tasks with task-defined phases for \textbf{CAP}, capturing phase-dependent corrective subspaces beyond normal-force compliance (e.g., tangential friction and torque cues for planar alignment and jamming relief). Table~\ref{tab:subspace} summarizes the phase-specific activated subspaces for \textbf{Fast}.

\textbf{(i) Charger Plug-in.}
Phases: \{\emph{approach, search, recovery, insert, done}\}.
\emph{Search} performs planar hole-alignment, where tangential friction forces and torques reveal misalignment and hole rim contact; thus corrections mainly lie in a planar subspace.
\emph{Recovery} indicates severe sticking or large wrench transients and requires an explicit retreat-and-retry intent switch, handled by the \textbf{Slow} planner rather than within-phase \textbf{Fast} micro-correction.
Notably, \emph{Recovery} may not occur in every episode; it is activated only when the above conditions are detected, while many successful trials proceed with \emph{search} followed by \emph{insert}.
\emph{Insert} emphasizes continuous advancement toward a fully seated insertion.

\textbf{(ii) USB Plug-in.}
Phases: same as \textbf{(i)}. In our setup, USB is more sensitive to small planar/yaw misalignment, often exhibiting friction-induced stick--slip and transient torques due to edge contact under tight tolerances.

\textbf{(iii) Drawer Opening.}
Phases: \{\emph{pick, unlock, pull, done}\}.
\emph{Unlock} overcomes initial stiction.
\emph{Pull} follows the drawer's constraint-guided sliding motion, where wrench feedback enforces directional compliance---driving along the opening direction while suppressing lateral forces that cause binding.

\textbf{(iv) Wiping (ID).}
Phases: \{\emph{pick, approach, wiping, done}\}.
Only \emph{wiping} is force-critical: vision is ambiguous about contact quality, as slight detachment yields ineffective wiping while over-pressing increases friction and induces oscillation; thus normal force feedback provides a direct signal for maintaining effective contact.

\textbf{(v) Wiping (OOD).}
Same phases as ID, but the board is raised by 3\,cm at test time while demonstrations are collected at the original height, creating an out-of-distribution contact condition.
This further tests whether a policy can leverage force feedback to compensate for unseen contact geometry and maintain stable wiping.

\begin{table}[t]
\centering
\small
\setlength{\tabcolsep}{6.5pt}
\renewcommand{\arraystretch}{0.95}
\caption{Corrective subspaces for force-critical phases across tasks}
\label{tab:subspace}
\begin{tabular*}{\columnwidth}{@{\extracolsep{\fill}}lccccc@{}}
\toprule
\textbf{subspace} & \textbf{search} & \textbf{insert} & \textbf{unlock} & \textbf{pull} & \textbf{wiping} \\
\midrule
$x$     & \cmark &  & \cmark & \cmark &  \\
$y$     & \cmark &  & \cmark & \cmark &  \\
$z$     &  & \cmark &  &  & \cmark \\
$\mathrm{roll}$    &  &  &  & \cmark &  \\
$\mathrm{pitch}$   &  &  &  & \cmark &  \\
$\mathrm{yaw}$     & \cmark & \cmark &  &  &  \\
\bottomrule
\end{tabular*}
\end{table}

\textbf{Metrics.}
For each method and each task, we run 20 evaluation trials with randomized initial conditions and report the success rate (SR).
For plug-in tasks, we regard partial insertion that does not reach a fully seated state as failure.
For wiping, we additionally evaluate contact quality and wiping effectiveness in Sec.~\ref{sec:wiping}.

\subsection{Baselines and Implementation}

\textbf{Baselines.}
We compare against four methods:
(i) \emph{Diffusion Policy (DP)}~\cite{chi2025diffusion}, a strong vision-only imitation learning policy;
(ii) \emph{DP (force-concat)}, which directly concatenates the force feature with the vision feature for action generation;
(iii) \emph{RDP}~\cite{xue2025reactive}, a representative slow--fast diffusion policy that leverages high-rate wrench/tactile feedback for within-chunk reactive execution; and
(iv) \emph{PhaForce (Ours)}.

\textbf{Implementation.}
For all methods, diffusion runs at $f_s=6$~Hz and we execute at a control rate of $f_c=24$~Hz with an action-chunk horizon $H_a=16$; we adopt latency matching by discarding the first few steps following UMI~\cite{chi2024universal} and RDP~\cite{xue2025reactive}, and send interpolated actions to the low-level controller at $>500$~Hz.
For \emph{PhaForce}, we use a wrench-history window of $H_w=36$ ($\approx 1.5\,s$) to capture short-term interaction dynamics.
Each RGB view is encoded into a 512-d embedding and concatenated into a visual token of dimension $d=1536$.
For \textbf{CAP}, we set $\delta=3$, $K_f=8$, and $\lambda_\phi=2$.
In \textbf{Slow}, we use multi-head cross-attention with $H=8$ heads (per-head dimension $d_k=d/H=192$).
For diffusion, we use a DDIM scheduler with $\epsilon$-prediction, using 100 timesteps at training and 10 timesteps at inference.
For the physical-prior teachers in \textbf{Fast}, we set $\alpha_x=\alpha_y=\alpha_z=5\times10^{-5}\,\mathrm{m/N}$ and $\alpha_{\mathrm{roll}}=\alpha_{\mathrm{pitch}}=\alpha_{\mathrm{yaw}}=3\times10^{-2}\,\mathrm{rad/(N\cdot m)}$. The average inference time per run is $\sim$120\,ms (Slow), $\sim$3\,ms (CAP), and $<\!1$\,ms (Fast).

\begin{figure*}[!t]
    \centering % 图片居中
    % 调整width=1\textwidth使图片铺满双栏宽度
    \includegraphics[width=1\textwidth]{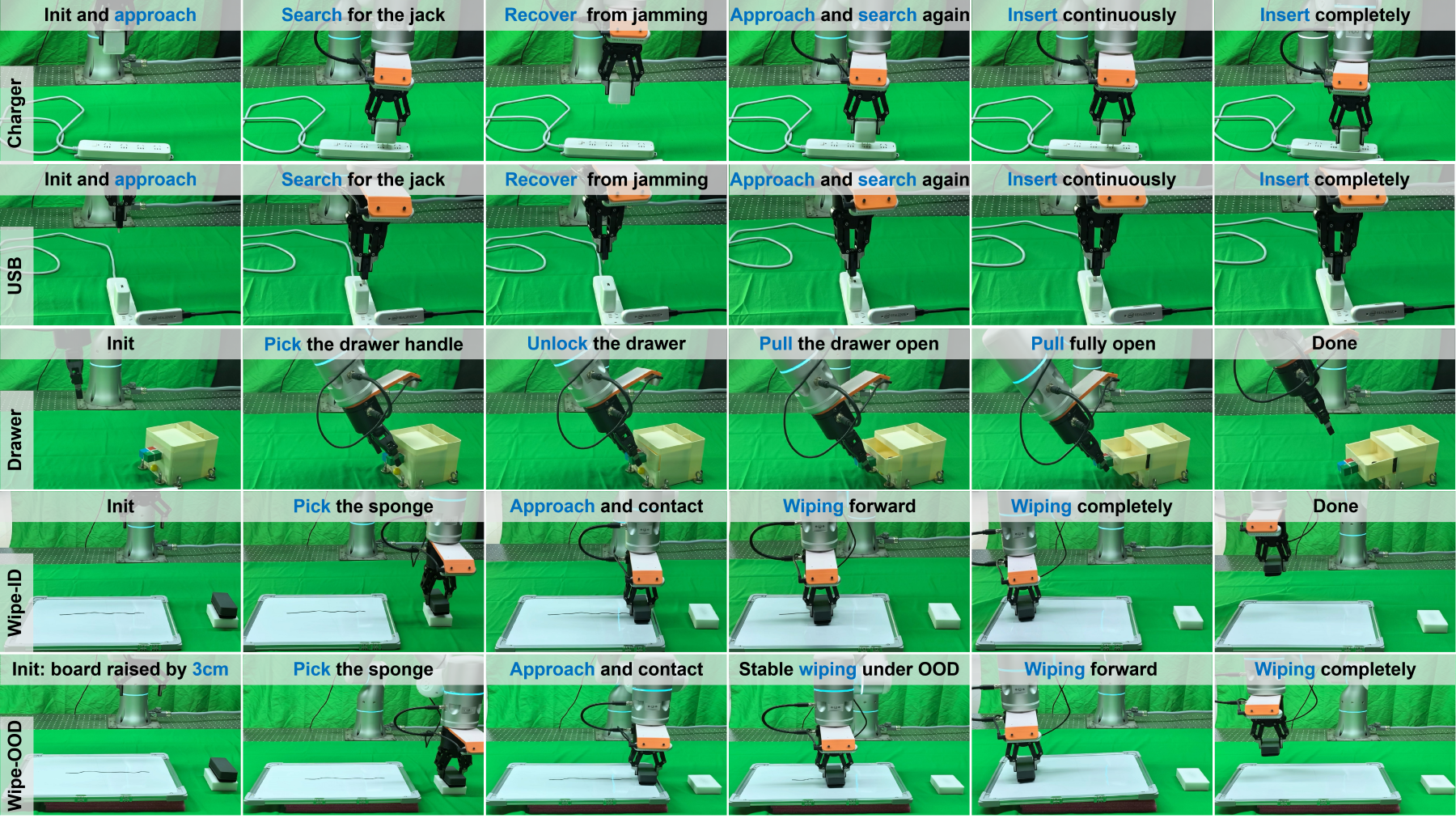}
    % 图片标题，label用于交叉引用
    \caption{We design five contact-rich tasks; each task exhibits \emph{varying contact states and phase belief}, and each phase activates different corrective subspaces. Each row illustrates the phase transitions in a task. PhaForce not only excels on in-distribution tasks but also remains stable under OOD shifts.}
    \label{fig:experiment}
\end{figure*}

% =========================
% IV-D Results & Analysis
% =========================
\subsection{Results and Analysis}
\label{sec:results_analysis}

\begin{table}[t]
\centering
\caption{Success rate (SR, \%) across different policies.}
\label{tab:main_results}
\footnotesize
\setlength{\tabcolsep}{3.1pt}
\renewcommand{\arraystretch}{1.12}

\begin{tabular}{l@{\hspace{2pt}}cccccc}
% \specialrule{thickness}{space_above}{space_below}
\specialrule{1.0pt}{0pt}{0.30em}  % top rule: add space below
\textbf{Method} & \textbf{Charger} & \textbf{USB} & \textbf{Drawer} & \textbf{Wipe-ID} & \textbf{Wipe-OOD} & \textbf{Avg} \\
\specialrule{0.5pt}{0.15em}{0.15em}  % mid rule: space above/below
DP                & 20 & 15 & 60 & \textbf{95} & 0  & 38 \\
DP (force-concat) & 20 & 20 & 50 & 85          & 0  & 35 \\
RDP               & 50 & 55 & 65 & 85          & 75 & 66 \\
PhaForce (ours)   & \textbf{80} & \textbf{85} & \textbf{85} & \textbf{95} & \textbf{85} & \textbf{86} \\
\specialrule{1.0pt}{0.20em}{0pt}  % bottom rule: add space above
\end{tabular}

\end{table}

Table~\ref{tab:main_results} reports success rates across five real-robot tasks.
Overall, \textbf{PhaForce} achieves the best (or tied-best) performance on all tasks.
Averaged over the three baselines, \textbf{PhaForce} improves the mean SR by \textbf{+40}~pp. 
Fig.~\ref{fig:experiment} further shows \textbf{PhaForce}'s execution over time, highlighting phase switches throughout each task.

\begin{figure}[t]
  \centering
  % 调整 trim={<left> <bottom> <right> <top>} 的数值来裁剪图片，单位是bp
  \includegraphics[width=1\columnwidth, trim={0 0 0 0}, clip]{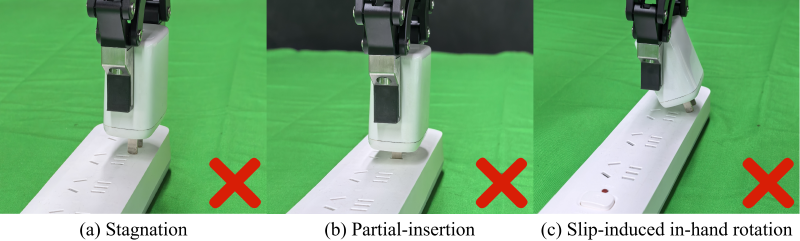}
    \caption{Three common baseline failure modes in plug-in tasks.}
  \label{fig:failure}
\end{figure}

\textbf{Plug-in tasks.}
Insertion is highly sensitive to small pose errors and local contact geometry, and typically involves phase switches such as planar search and recovery.
As shown in Fig.~\ref{fig:failure}, we observe three common failure modes of baselines:
\emph{(i) Stagnation:} after a slight misalignment, the end-effector gets stuck at the socket entrance and fails to trigger planar search or retreat-and-retry.
\emph{(ii) Partial insertion:} the plug enters the socket but remains not fully seated, resulting in an incomplete insertion.
\emph{(iii) Slip-induced in-hand rotation:} rim collisions with excessive contact force can cause the connector to slip inside the gripper and rotate substantially, misorienting the plug.
The last mode is an \emph{unintended} disturbance rather than an intended search subspace, and is difficult to compensate without demonstrations covering large reorientation.

\begin{figure}[t]
  \centering
  % 调整 trim={<left> <bottom> <right> <top>} 的数值来裁剪图片，单位是bp
  \includegraphics[width=1\columnwidth, trim={0 0 0 0}, clip]{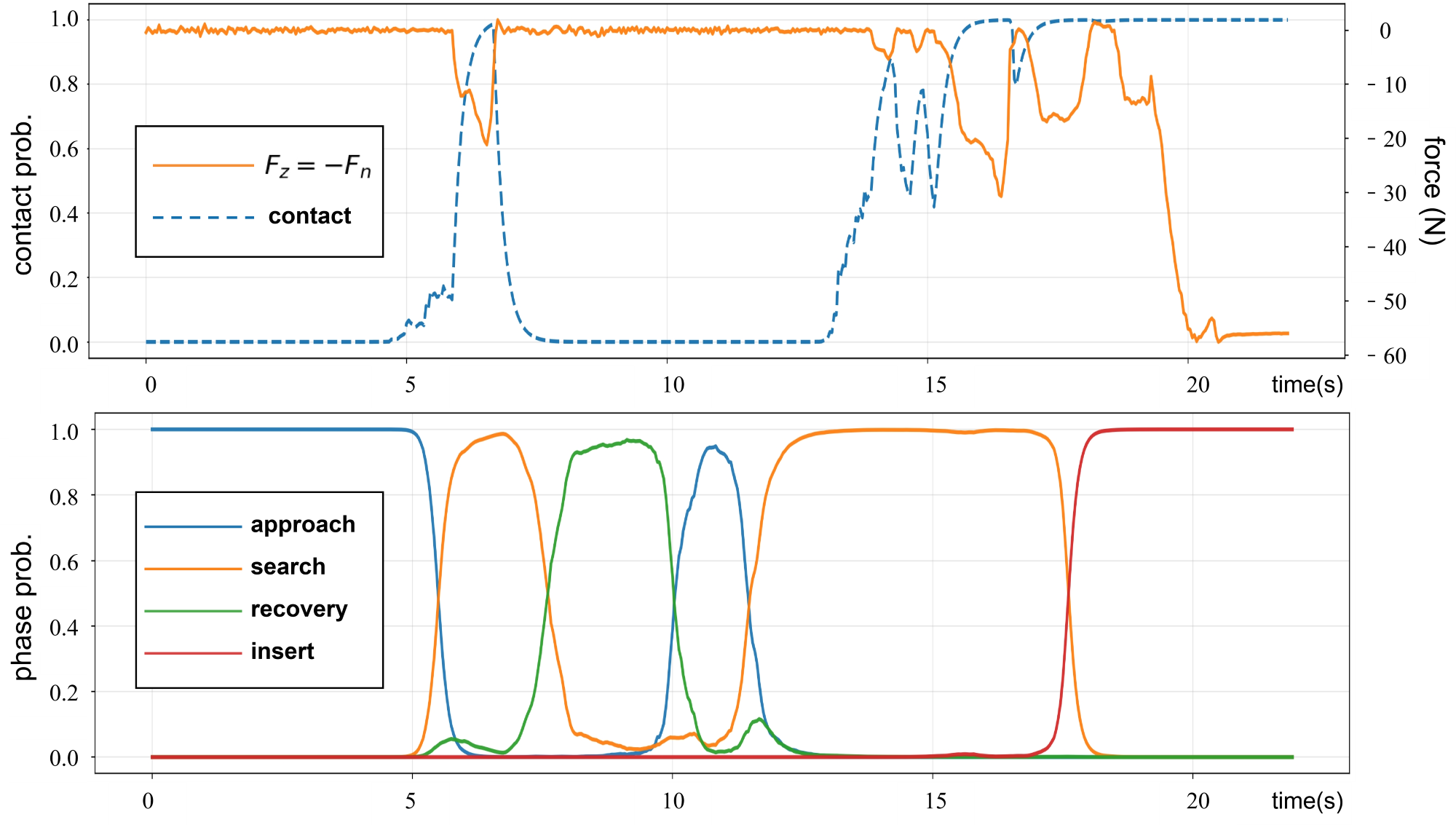}
    \caption{
    We visualize the contact probability, phase belief, and $z$-axis force $F_z$ in a USB Plug-in task (curves are smoothed for visualization).
    }
  \label{fig:force}
\end{figure}

\emph{DP} often fails to achieve millimeter-level alignment because it must infer the contact state and corrective direction purely from images.
\emph{DP (force-concat)} yields only limited gains because naive wrench concatenation lacks an explicit mechanism to convert high-frequency force feedback into timely within-chunk micro-corrections~\cite{xue2025reactive}.

\emph{RDP} improves performance via fast closed-loop refinement, yet it still underperforms \textbf{PhaForce} on fine insertion.
We conjecture that, for such millimeter-sensitive tasks, latent action-space planning can degrade execution precision for contact alignment, consistent with the precision-loss effects attributed to latent compression in~\cite{chen2025implicitrdp}. In contrast, \textbf{PhaForce}'s \textbf{Slow} predicts executable actions directly in the \emph{real} action space, rather than latent space.
Moreover, without an explicit phase belief, RDP may struggle to reliably switch into \emph{recovery}, making it harder to escape stagnation at the socket entrance.

\textbf{PhaForce} mitigates these issues by using phase belief to condition and gate search/recovery behaviors, enabling timely retreat-and-retry instead of stagnation. As shown in Fig.~\ref{fig:force}, CAP's anticipatory contact/phase predictions align with wrench transients. Moreover, \textbf{PhaForce} applies targeted within-chunk residual corrections in task-relevant subspaces while preserving the vision-conditioned semantics of the slow planning chunk. It also alleviates partial-insertion failures where baselines engage the hole yet fail to achieve a fully seated insertion.
For slip events, we reduce their occurrence by avoiding excessive contact and by triggering recovery once abnormal wrench signals are detected; handling large in-hand rotations more fundamentally remains future work (e.g., explicit in-hand pose estimation or enriched demonstrations).

\textbf{Drawer task.}
For Drawer Opening, \textbf{PhaForce} achieves a consistent improvement and we attribute this gain to phase-aware force utilization that facilitates compliant pulling (with small roll/pitch compliance) under friction variations and occasional binding, enabling timely adjustments rather than persisting with a misaligned pull.

% =========================
% IV-E Wiping: Contact Quality & Effectiveness
% =========================
\subsection{Wiping: Contact Quality and Effectiveness}
\label{sec:wiping}

Success rate alone is insufficient for wiping, since a policy may complete the motion while applying excessive force or experiencing frequent contact dropouts.
For both ID and OOD, we report a wiping score (1 for fully wiped, 0.5 for partial wiping, and 0 for no erasure), mean contact normal force $\overline{F_n}$, and the over-pressure ($F_n>25$~N) / under-pressure ($F_n<2.5$~N) time ratios.
For reference, in our demonstrations the mean normal force during contact is $18.7$~N; we empirically set the target normal force to $F_n^{\ast}=12$~N, corresponding to $F_z^{\ast}=-12$~N under the convention $F_z=-F_n$, to reduce over-pressure while maintaining reliable erasure.

\begin{table}[t]
\centering
\caption{Wiping (ID): deeper evaluation beyond SR.}
\label{tab:wipe_id}
\footnotesize
\renewcommand{\arraystretch}{1.12}
\begin{tabular*}{\columnwidth}{@{\extracolsep{\fill}}lccccc}
\toprule
\textbf{Method} & \textbf{SR} & \textbf{Score} & $\boldsymbol{\overline{F_n}}$\textbf{ (N)} & \textbf{Over} & \textbf{Under} \\
\midrule
DP                  & \textbf{95} & 0.80 & 17.3 & 21.5\% & 7.4\% \\
DP (force-concat)   & 85          & 0.75 & 17.0 & 16.8\% & 5.6\% \\
RDP                 & 85          & 0.65 & 13.6 & 5.7\%  & 2.3\% \\
PhaForce w/o Fast   & \textbf{95} & 0.70 & 15.1  & 14.3\% & 5.0\% \\
PhaForce (ours)     & \textbf{95} & \textbf{0.85} & 12.3 & \textbf{4.5\%} & \textbf{2.0\%} \\
\bottomrule
\end{tabular*}
\end{table}

\begin{table}[t]
\centering
\caption{Wiping (OOD): deeper evaluation beyond SR (\texttt{-} denotes a metric that is not applicable when SR=0).}
\label{tab:wipe_ood}
\footnotesize
\renewcommand{\arraystretch}{1.12}
\begin{tabular*}{\columnwidth}{@{\extracolsep{\fill}}lccccc}
\toprule
\textbf{Method} & \textbf{SR} & \textbf{Score} & $\boldsymbol{\overline{F_n}}$\textbf{ (N)} & \textbf{Over} & \textbf{Under} \\
\midrule
DP                  & 0  & --  & 46.2 & 85.3\% & -- \\
DP (force-concat)   & 0  & --  & 44.6 & 86.7\% & -- \\
RDP                 & 75 & 0.65 & 9.2  & 7.2\%  & \textbf{0.22\%} \\
PhaForce w/o Fast   & 0  & --  & 48.3 & 80.3\% & -- \\
PhaForce (ours)     & \textbf{85} & \textbf{0.75} & 14.3 & \textbf{7.0\%} & 0.34\% \\
\bottomrule
\end{tabular*}
\end{table}

\textbf{Wiping (ID).}
As shown in Table~\ref{tab:wipe_id}, \textbf{PhaForce} achieves the best overall wiping outcome, with the highest score, the lowest under-/over-pressure ratios.
By contrast, \emph{DP} attains strong SR/Score, but its contact quality is unstable, with frequent over-pressure and contact dropouts, consistent with small teleoperation jitter in demonstrations manifesting as force fluctuations during contact.
\emph{DP (force-concat)} and \emph{RDP} occasionally fail to grasp the sponge, highlighting the risk of using wrench feedback without phase scheduling: in non-contact stages, wrench signals are often noise-dominated, thereby hurting overall performance~\cite{he2025foar}.
Meanwhile, \emph{RDP} markedly reduces unstable contact events, yet yields a lower wiping score, plausibly because latent action-space planning can degrade fine-grained visual localization needed for precise erasing of the notes.
Overall, \textbf{PhaForce} benefits from (i) \textbf{CAP} for unified phase-aware scheduling of force usage, (ii) \textbf{Slow} visual-force fusion via \emph{ORI} that preserves vision-dominant task semantics, and (iii) the \textbf{Fast} corrector regulates the contact normal force via force feedback in the corrective subspace.

\textbf{Wiping (OOD).}
As shown in Table~\ref{tab:wipe_ood}, chunk-level diffusion planners without fast correction fail completely (\emph{DP}, \emph{DP (force-concat)}, and \emph{PhaForce w/o Fast}), exhibiting sustained over-pressure and zero success.
This collapse is mainly due to the height mismatch: the policy overfits to the demonstration height, causing the end-effector to either over-press and stall or stick to the board and drag quasi-statically under large friction, thus failing to execute the intended wiping motion.
In contrast, \emph{RDP} remains feasible under OOD, highlighting the advantage of a slow--fast design for contact adaptation, and \textbf{PhaForce} further improves wiping score with comparable contact stability, showing that \textbf{Fast} is key to compensating the OOD height mismatch.

% =========================
% IV-F Ablations
% =========================
\subsection{Ablations}
\label{sec:ablation}

To validate the effectiveness of key components of \textbf{PhaForce}, we conduct ablations on two representative tasks: \emph{USB Plug-in}, a multi-phase task with distinct phase transitions, and \emph{Wiping (OOD)}, which requires robust contact adaptation under environment shifts.
Specifically, we consider three ablation variants:
(i) \emph{PhaForce (w/o PB)}, removing phase belief by replacing $\mathbf{p}_t$ with a uniform prior over $K$ phases during \emph{both} training and testing, i.e., $p_t^{(k)} \equiv 1/K$;
(ii) \emph{PhaForce (w/o ORI)}, replacing the fused token $v_t'$ with the cross-attention output $\Delta_t$ as diffusion conditioning;
and (iii) \emph{PhaForce (w/o Fast)}, removing the \textbf{Fast} residual corrector.

\begin{table}[t]
\centering
\caption{Ablation results over 20 real-robot trials per method.}
\label{tab:ablation}
\footnotesize
\renewcommand{\arraystretch}{1.12}
\begin{tabular*}{\columnwidth}{@{\extracolsep{\fill}}lccc}
\toprule
\textbf{Method} & \textbf{USB SR} & \textbf{Wipe-OOD SR} & \textbf{Wiping Score} \\
\midrule
PhaForce (w/o PB)     & 25 & 45 & 0.60 \\
PhaForce (w/o ORI)    & 35 & 60 & 0.45 \\
PhaForce (w/o Fast)   & 50 & 0  & -- \\
PhaForce (ours)       & \textbf{85} & \textbf{85} & \textbf{0.75} \\
\bottomrule
\end{tabular*}
\end{table}

\textbf{Results.}
Table~\ref{tab:ablation} shows that removing phase belief (\emph{w/o PB}) severely hurts \emph{USB Plug-in} (SR 85$\rightarrow$25), indicating that explicit phase scheduling is essential to \emph{route} corrections to the right subspaces and \emph{trigger} timely search/recovery/insert transitions in a multi-phase insertion.
Replacing \emph{ORI} with direct cross-attention conditioning (\emph{w/o ORI}) degrades both SR and wiping score on \emph{Wipe-OOD} (SR 85$\rightarrow$60; score 0.75$\rightarrow$0.45), supporting that \emph{ORI} preserves vision-dominant semantics while injecting force information.
Finally, removing \textbf{Fast} (\emph{w/o Fast}) collapses \emph{Wipe-OOD} (SR 0), confirming that high-rate residual correction is indispensable for stabilizing contact (rapidly compensating F/T transients) when the environment deviates from the demonstrations.

\section{CONCLUSIONS}

In this paper, we propose \textbf{PhaForce}, a phase-scheduled visual--force policy learning framework that integrates low-rate generative planning with high-rate reactive correction for contact-rich manipulation.
PhaForce combines a contact-aware phase predictor (\textbf{CAP}) that delivers a global contact/phase schedule, a \textbf{Slow} diffusion planner that performs dual-gated vision--force fusion with orthogonal residual injection, and a \textbf{Fast} residual corrector that performs within-chunk, subspace-specific closed-loop refinement.
Real-robot experiments across five tasks show that PhaForce consistently outperforms strong baselines in both ID and OOD settings, excelling in both phase-transition-intensive skills and sustained-contact skills beyond success rate alone.
Future work will explore learning the Fast corrector with reinforcement learning beyond supervised residual targets, and extend PhaForce from single-task imitation to VLA models that generalize across diverse skills and embodiments.


%%%%%%%%%%%%%%%%%%%%%%%%%%%%%%%%%%%%%%%%%%%%%%%%%%%%%%%%%%%%%%%%%%%%%%%%%%%%%%%%

%%%%%%%%%%%%%%%%%%%%%%%%%%%%%%%%%%%%%%%%%%%%%%%%%%%%%%%%%%%%%%%%%%%%%%%%%%%%%%%%

%%%%%%%%%%%%%%%%%%%%%%%%%%%%%%%%%%%%%%%%%%%%%%%%%%%%%%%%%%%%%%%%%%%%%%%%%%%%%%%%

\bibliographystyle{IEEEtran}
\bibliography{references}

@inproceedings{stepputtis2022system,
  title={A system for imitation learning of contact-rich bimanual manipulation policies},
  author={Stepputtis, Simon and Bandari, Maryam and Schaal, Stefan and Amor, Heni Ben},
  booktitle={2022 IEEE/RSJ International Conference on Intelligent Robots and Systems (IROS)},
  pages={11810--11817},
  year={2022},
  organization={IEEE}
}

@inproceedings{hou2025adaptive,
  title={Adaptive compliance policy: Learning approximate compliance for diffusion guided control},
  author={Hou, Yifan and Liu, Zeyi and Chi, Cheng and Cousineau, Eric and Kuppuswamy, Naveen and Feng, Siyuan and Burchfiel, Benjamin and Song, Shuran},
  booktitle={2025 IEEE International Conference on Robotics and Automation (ICRA)},
  pages={4829--4836},
  year={2025},
  organization={IEEE}
}

@inproceedings{li2025adaptive,
  title={Adaptive visuo-tactile fusion with predictive force attention for dexterous manipulation},
  author={Li, Jinzhou and Wu, Tianhao and Zhang, Jiyao and Chen, Zeyuan and Jin, Haotian and Wu, Mingdong and Shen, Yujun and Yang, Yaodong and Dong, Hao},
  booktitle={2025 IEEE/RSJ International Conference on Intelligent Robots and Systems (IROS)},
  pages={3232--3239},
  year={2025},
  organization={IEEE}
}

@article{tsuji2024adaptive,
  title={Adaptive contact-rich manipulation through few-shot imitation learning with Force-Torque feedback and pre-trained object representations},
  author={Tsuji, Chikaha and Coronado, Enrique and Osorio, Pablo and Venture, Gentiane},
  journal={IEEE Robotics and Automation Letters},
  volume={10},
  number={1},
  pages={240--247},
  year={2024},
  publisher={IEEE}
}

@article{zhou2025admittance,
  title={Admittance visuomotor policy learning for general-purpose contact-rich manipulations},
  author={Zhou, Bo and Jiao, Ruixuan and Li, Yi and Yuan, Xiaogang and Fang, Fang and Li, Shihua},
  journal={IEEE Transactions on Industrial Electronics},
  year={2025},
  publisher={IEEE}
}

@article{chen2025dexforce,
  title={Dexforce: Extracting force-informed actions from kinesthetic demonstrations for dexterous manipulation},
  author={Chen, Claire and Yu, Zhongchun and Choi, Hojung and Cutkosky, Mark and Bohg, Jeannette},
  journal={IEEE Robotics and Automation Letters},
  year={2025},
  publisher={IEEE}
}

@article{zhang2026dextac,
  title={DexTac: Learning Contact-aware Visuotactile Policies via Hand-by-hand Teaching},
  author={Zhang, Xingyu and Zhang, Chaofan and Zhang, Boyue and Peng, Zhinan and Cui, Shaowei and Wang, Shuo},
  journal={arXiv preprint arXiv:2601.21474},
  year={2026}
}

@article{ge2025filic,
  title={FILIC: Dual-Loop Force-Guided Imitation Learning with Impedance Torque Control for Contact-Rich Manipulation Tasks},
  author={Ge, Haizhou and Jia, Yufei and Li, Zheng and Li, Yue and Chen, Zhixing and Huang, Ruqi and Zhou, Guyue},
  journal={arXiv preprint arXiv:2509.17053},
  year={2025}
}

@article{li2025flow,
  title={Flow with the Force Field: Learning 3D Compliant Flow Matching Policies from Force and Demonstration-Guided Simulation Data},
  author={Li, Tianyu and Li, Yihan and Zhang, Zizhe and Figueroa, Nadia},
  journal={arXiv preprint arXiv:2510.02738},
  year={2025}
}

@article{he2025foar,
  title={Foar: Force-aware reactive policy for contact-rich robotic manipulation},
  author={He, Zihao and Fang, Hongjie and Chen, Jingjing and Fang, Hao-Shu and Lu, Cewu},
  journal={IEEE Robotics and Automation Letters},
  year={2025},
  publisher={IEEE}
}

@inproceedings{liu2025forcemimic,
  title={Forcemimic: Force-centric imitation learning with force-motion capture system for contact-rich manipulation},
  author={Liu, Wenhai and Wang, Junbo and Wang, Yiming and Wang, Weiming and Lu, Cewu},
  booktitle={2025 IEEE International Conference on Robotics and Automation (ICRA)},
  pages={1105--1112},
  year={2025},
  organization={IEEE}
}

@article{chen2025implicitrdp,
  title={ImplicitRDP: An End-to-End Visual-Force Diffusion Policy with Structural Slow-Fast Learning},
  author={Chen, Wendi and Xue, Han and Wang, Yi and Zhou, Fangyuan and Lv, Jun and Jin, Yang and Tang, Shirun and Wen, Chuan and Lu, Cewu},
  journal={arXiv preprint arXiv:2512.10946},
  year={2025}
}

@article{choi2026wild,
  title={In-the-Wild Compliant Manipulation with UMI-FT},
  author={Choi, Hojung and Hou, Yifan and Pan, Chuer and Hong, Seongheon and Patel, Austin and Xu, Xiaomeng and Cutkosky, Mark R and Song, Shuran},
  journal={arXiv preprint arXiv:2601.09988},
  year={2026}
}

@article{aburub2024learning,
  title={Learning diffusion policies from demonstrations for compliant contact-rich manipulation},
  author={Aburub, Malek and Beltran-Hernandez, Cristian C and Kamijo, Tatsuya and Hamaya, Masashi},
  journal={arXiv preprint arXiv:2410.19235},
  year={2024}
}

@inproceedings{kamijo2024learning,
  title={Learning variable compliance control from a few demonstrations for bimanual robot with haptic feedback teleoperation system},
  author={Kamijo, Tatsuya and Beltran-Hernandez, Cristian C and Hamaya, Masashi},
  booktitle={2024 IEEE/RSJ International Conference on Intelligent Robots and Systems (IROS)},
  pages={12663--12670},
  year={2024},
  organization={IEEE}
}

@article{lee2025manipforce,
  title={ManipForce: Force-Guided Policy Learning with Frequency-Aware Representation for Contact-Rich Manipulation},
  author={Lee, Geonhyup and Lee, Yeongjin and Kim, Kangmin and Lee, Seongju and Noh, Sangjun and Back, Seunghyeok and Lee, Kyoobin},
  journal={arXiv preprint arXiv:2509.19047},
  year={2025}
}

@article{xue2025reactive,
  title={Reactive diffusion policy: Slow-fast visual-tactile policy learning for contact-rich manipulation},
  author={Xue, Han and Ren, Jieji and Chen, Wendi and Zhang, Gu and Fang, Yuan and Gu, Guoying and Xu, Huazhe and Lu, Cewu},
  journal={arXiv preprint arXiv:2503.02881},
  year={2025}
}

@article{kang2025robotic,
  title={Robotic compliant object prying using diffusion policy guided by vision and force observations},
  author={Kang, Jeon Ho and Joshi, Sagar and Huang, Ruopeng and Gupta, Satyandra K},
  journal={IEEE Robotics and Automation Letters},
  year={2025},
  publisher={IEEE}
}

@inproceedings{wu2025tacdiffusion,
  title={Tacdiffusion: Force-domain diffusion policy for precise tactile manipulation},
  author={Wu, Yansong and Chen, Zongxie and Wu, Fan and Chen, Lingyun and Zhang, Liding and Bing, Zhenshan and Swikir, Abdalla and Haddadin, Sami and Knoll, Alois},
  booktitle={2025 IEEE International Conference on Robotics and Automation (ICRA)},
  pages={11831--11837},
  year={2025},
  organization={IEEE}
}

@article{sun2024force,
  title={Force-constrained visual policy: Safe robot-assisted dressing via multi-modal sensing},
  author={Sun, Zhanyi and Wang, Yufei and Held, David and Erickson, Zackory},
  journal={IEEE Robotics and Automation Letters},
  volume={9},
  number={5},
  pages={4178--4185},
  year={2024},
  publisher={IEEE}
}

@article{yu2025forcevla,
  title={Forcevla: Enhancing vla models with a force-aware moe for contact-rich manipulation},
  author={Yu, Jiawen and Liu, Hairuo and Yu, Qiaojun and Ren, Jieji and Hao, Ce and Ding, Haitong and Huang, Guangyu and Huang, Guofan and Song, Yan and Cai, Panpan and others},
  journal={arXiv preprint arXiv:2505.22159},
  year={2025}
}

@article{zhang2025ta,
  title={Ta-vla: Elucidating the design space of torque-aware vision-language-action models},
  author={Zhang, Zongzheng and Xu, Haobo and Yang, Zhuo and Yue, Chenghao and Lin, Zehao and Gao, Huan-ang and Wang, Ziwei and Zhao, Hao},
  journal={arXiv preprint arXiv:2509.07962},
  year={2025}
}

@article{chi2025diffusion,
  title={Diffusion policy: Visuomotor policy learning via action diffusion},
  author={Chi, Cheng and Xu, Zhenjia and Feng, Siyuan and Cousineau, Eric and Du, Yilun and Burchfiel, Benjamin and Tedrake, Russ and Song, Shuran},
  journal={The International Journal of Robotics Research},
  volume={44},
  number={10-11},
  pages={1684--1704},
  year={2025},
  publisher={Sage Publications Sage UK: London, England}
}

@article{zhao2023learning,
  title={Learning fine-grained bimanual manipulation with low-cost hardware},
  author={Zhao, Tony Z and Kumar, Vikash and Levine, Sergey and Finn, Chelsea},
  journal={arXiv preprint arXiv:2304.13705},
  year={2023}
}

@article{chi2024universal,
  title={Universal manipulation interface: In-the-wild robot teaching without in-the-wild robots},
  author={Chi, Cheng and Xu, Zhenjia and Pan, Chuer and Cousineau, Eric and Burchfiel, Benjamin and Feng, Siyuan and Tedrake, Russ and Song, Shuran},
  journal={arXiv preprint arXiv:2402.10329},
  year={2024}
}

@article{liu2024rdt,
  title={Rdt-1b: a diffusion foundation model for bimanual manipulation},
  author={Liu, Songming and Wu, Lingxuan and Li, Bangguo and Tan, Hengkai and Chen, Huayu and Wang, Zhengyi and Xu, Ke and Su, Hang and Zhu, Jun},
  journal={arXiv preprint arXiv:2410.07864},
  year={2024}
}

@article{kim2024openvla,
  title={Openvla: An open-source vision-language-action model},
  author={Kim, Moo Jin and Pertsch, Karl and Karamcheti, Siddharth and Xiao, Ted and Balakrishna, Ashwin and Nair, Suraj and Rafailov, Rafael and Foster, Ethan and Lam, Grace and Sanketi, Pannag and others},
  journal={arXiv preprint arXiv:2406.09246},
  year={2024}
}

@article{intelligence2025pi_,
  title={{$\pi_{0.5}$}: {A} {Vision-Language-Action} {Model} with {Open-World} {Generalization}},
  author={{Physical Intelligence} and Black, Kevin and Brown, Noah and Darpinian, James and Dhabalia, Karan and Driess, Danny and Esmail, Adnan and Equi, Michael and Finn, Chelsea and Fusai, Niccolo and others},
  journal={arXiv preprint arXiv:2504.16054},
  year={2025}
}

@article{black2024pi_0,
  title={{$\pi_0$}: {A} {Vision-Language-Action} {Flow} {Model} for {General} {Robot} {Control}},
  author={Black, Kevin and Brown, Noah and Driess, Danny and Esmail, Adnan and Equi, Michael and Finn, Chelsea and Fusai, Niccolo and Groom, Lachy and Hausman, Karol and Ichter, Brian and others},
  journal={arXiv preprint arXiv:2410.24164},
  year={2024}
}

@article{ho2020denoising,
  title={Denoising diffusion probabilistic models},
  author={Ho, Jonathan and Jain, Ajay and Abbeel, Pieter},
  journal={Advances in neural information processing systems},
  volume={33},
  pages={6840--6851},
  year={2020}
}

@article{song2020denoising,
  title={Denoising diffusion implicit models},
  author={Song, Jiaming and Meng, Chenlin and Ermon, Stefano},
  journal={arXiv preprint arXiv:2010.02502},
  year={2020}
}

\end{document}